\documentclass[letterpaper, 10 pt, conference]{ieeeconf}  % Comment this line out
\IEEEoverridecommandlockouts                              % This command is only
\usepackage[utf8]{inputenc}
\usepackage[T1]{fontenc}
\usepackage{amssymb}
\usepackage{amsmath}
\usepackage[ruled]{algorithm2e}
\usepackage{amsfonts}
\usepackage{dsfont}
\usepackage{graphicx}   
\usepackage{url}   
\usepackage[backend=biber]{biblatex}
\usepackage{caption}
\usepackage{subcaption} 
\usepackage{stfloats}

\usepackage{xcolor,colortbl}

\definecolor{Gray}{gray}{0.85}
\definecolor{LightCyan}{rgb}{0.88,1,1}

\newcolumntype{a}{>{\columncolor{Gray}}c}
\newcolumntype{b}{>{\columncolor{white}}c}

\title{\LARGE \bf
HOME: Heatmap Output for future Motion Estimation
}

\author{Thomas Gilles$^{1, 2}$, Stefano Sabatini$^{1}$, Dzmitry Tsishkou$^{1}$, Bogdan Stanciulescu$^{2}$, Fabien Moutarde$^{2}$% <-this % stops a space

\thanks{$^{1}$IoV team, Paris Research Center, Huawei Technologies France}%
\thanks{$^{2}$MINES ParisTech, PSL University, Center for robotics}%
\thanks{Contact: thomas.gilles@mines-paristech.fr}%
}

\begin{document}

\maketitle
\thispagestyle{empty}
\pagestyle{empty}

%%%%%%%%%%%%%%%%%%%%%%%%%%%%%%%%%%%%%%%%%%%%%%%%%%%%%%%%%%%%%%%%%%%%%%%%%%%%%%%%
\begin{abstract}

In this paper, we propose HOME, a framework tackling the motion forecasting problem with an image output representing the probability distribution of the agent's future location. This method allows for a simple architecture with classic convolution networks coupled with attention mechanism for agent interactions, and outputs an unconstrained 2D top-view representation of the agent's possible future. Based on this output, we design two methods to sample a finite set of agent's future locations. These methods allow us to control the optimization trade-off between miss rate and final displacement error for multiple modalities without having to retrain any part of the model. We apply our method to the Argoverse Motion Forecasting Benchmark and achieve 1$^{st}$ place on the online leaderboard.

\end{abstract}

%%%%%%%%%%%%%%%%%%%%%%%%%%%%%%%%%%%%%%%%%%%%%%%%%%%%%%%%%%%%%%%%%%%%%%%%%%%%%%%%
\section{INTRODUCTION}

Forecasting the future motion of surrounding actors is an essential part of the autonomous driving pipeline, necessary for safe planning and useful for simulation of realistic behaviors. In order to capture the complexity of a driving scenario, the prediction model needs to take into account the local map, the past trajectory of the predicted agent and the interactions with other actors. Its output needs to be multimodal to cover the different choices a driver could make, between going straight or turning, slowing down or overtaking. Each modality proposed should represent a possible trajectory that an agent could take in the immediate future.

The challenge in motion prediction resides not in having the absolute closest trajectory to the ground truth, but rather in avoiding big failures where a possibility has not been considered, and the future is totally missed by all modalities. An accident will rarely happen because most predictions are offset by half a meter, but rather because of one single case where a lack of coverage led to a miss of more than a few meters.

A classic way to obtain $k$ modalities is to design a model that outputs a fixed number of $k$ future trajectories \cite{cui2019multimodal, mercat2020multi, messaoud2020multi, liang2020learning}, as a regression problem. This approach has however significant drawbacks, as training predictions all together leads to mode collapse. The common solution to this problem is to only train the closest prediction to the ground truth, but this diminishes the training data allocated to each predicted modality as only one is learning at each sample. 

Later methods adapt the model to the multi-modal problem by conditioning the prediction to specific inputs such as lanes \cite{khandelwal2020if} or targets \cite{zhao2020tnt}. Finally, recent methods use the topological lane graph itself to generate trajectory for each node \cite{zeng2021lanercnn}. However each of these model constrains its prediction space to a restricted representation, that may be limited to represent the actual diversity of possible futures. For example, if the predicted modalities are constrained to the High Definition map graph, it becomes very hard to predict agent breaking traffic rules or slowing down to park at the side of the road.

\begin{figure}[t]
\centerline{\includegraphics[width=0.9\columnwidth]{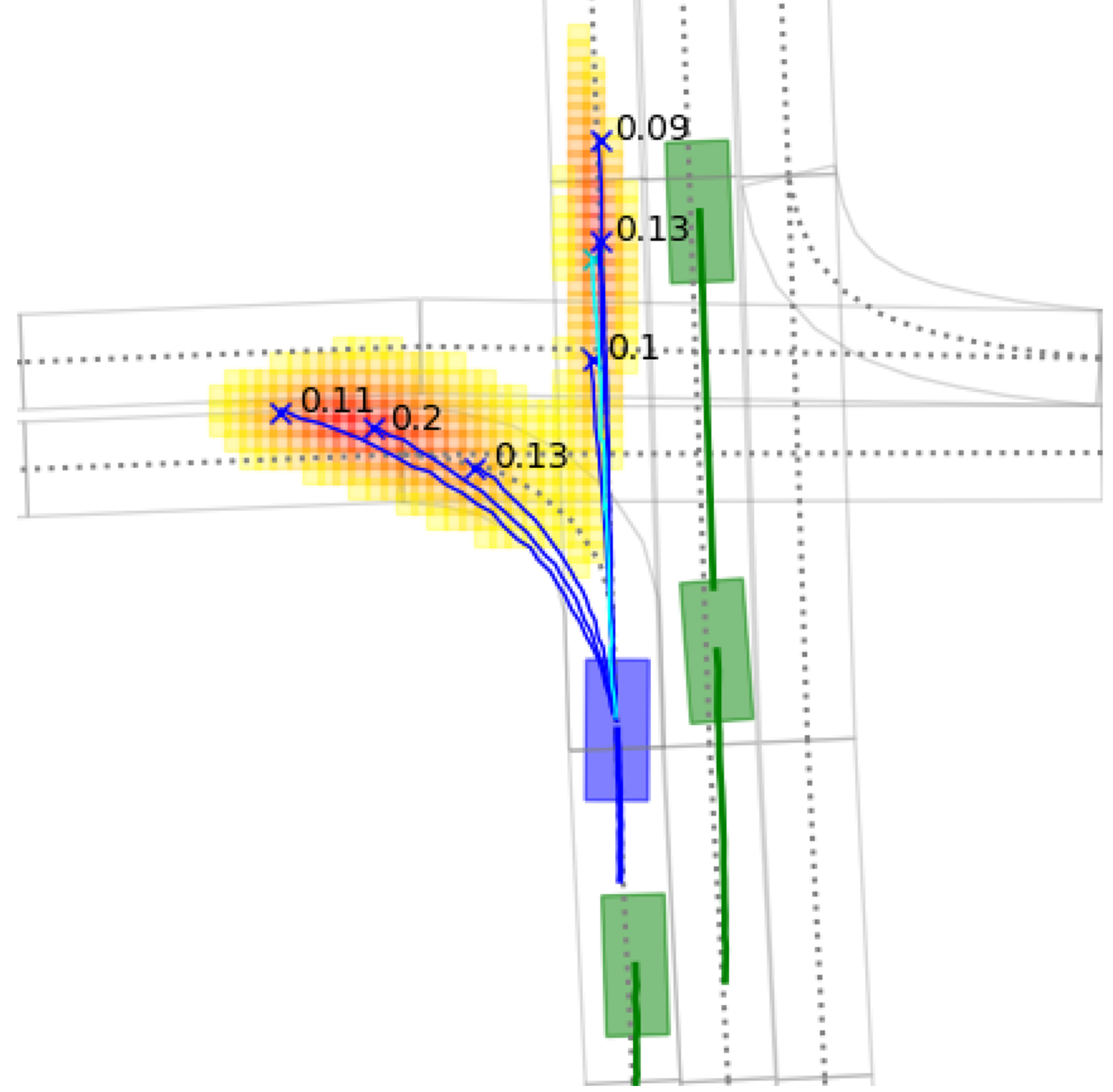}}
\caption{Summary of our approach. The yellow/red heatmap is our predicted probability distribution and the blue points are the sampled final point predictions. }
\label{fig:approach}
\end{figure}

\begin{figure*}[t]
\centerline{\includegraphics[width=2\columnwidth]{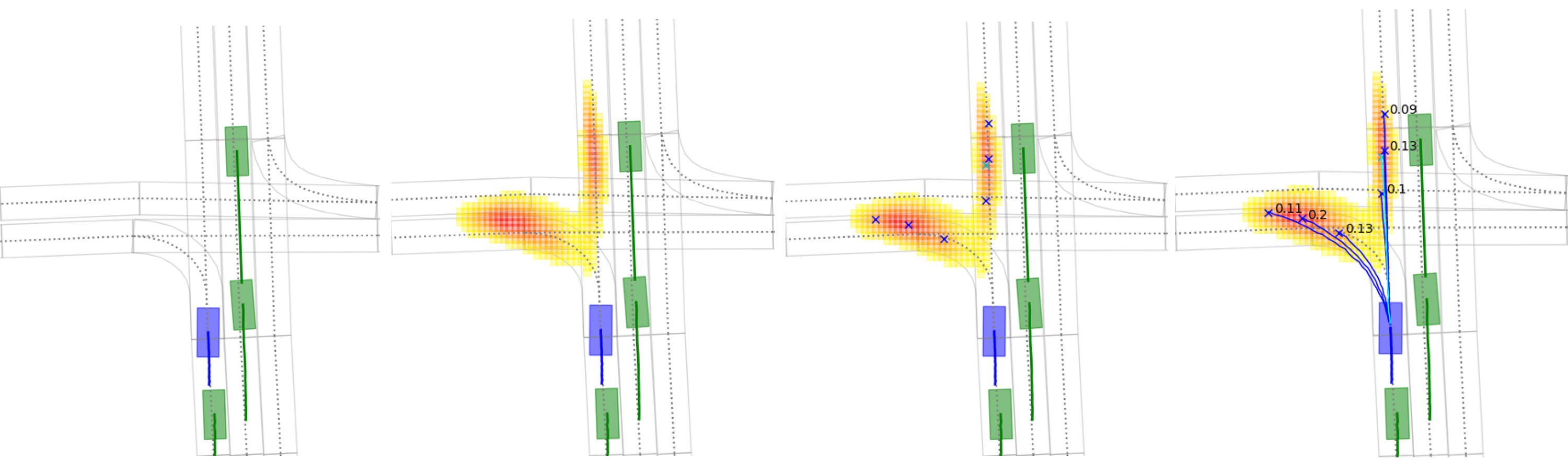}}
\caption{HOME pipeline. a) Context map, target agent (blue) and neighbor (green) trajectories are given as input to the network. b) Heatmap output of the network. c) Sampled final points. d) Trajectories are built for each final point}
\label{fig:pipeline}
\end{figure*}

In this paper, following the same principle as recent state of the art method, which is that a future trajectory can be almost fully defined by its final point \cite{zhao2020tnt, zeng2021lanercnn}, we reformulate the prediction problem in three steps. We first represent the possible futures distribution by a 2D probability heatmap that gives an unconstrained approximation of the probability of the agent position. This heatmap is represented as a squared image and it naturally accommodates for multimodal predictions where each pixel represent a possible future position of the target agent. It also enables to fully describe the future uncertainty in a probability distribution, without having to choose its modes or means. In a second step, we sample from the heatmap a finite number of possible future locations with the possibility to choose which metric we want to optimize without retraining the model. Finally, we build the full trajectories based on the past history and conditioned on the sampled final points.

Our contributions are summarized as follow:
\begin{itemize}
    \item We present a simple model architecture made of a convolutional neural network (CNN), a recurrent neural network (RNN) and an Attention module, with a heatmap output allowing for easy and efficient training.
    \item We design two  sampling algorithms from this heatmap output, optimizing MR$_k$ or minFDE$_k$ respectively.
    \item We highlight a trade-off between both metrics, and show that our sampling algorithm allows us to control this trade-off with a simple parameter.
\end{itemize}

\section{RELATED WORK}

Deep learning has brought great progress to the motion forecasting results \cite{mozaffari2020deep}. A classic CNN architecture can be applied to a rasterized map to predict 2D coordinates \cite{cui2019multimodal}.

In order to model interactions better between driving agents, attention has been introduced in multiple methods. The approach of \cite{mercat2020multi} encodes separately agents and centerlines with 1D CNN and LSTM and then applies multi-head attention from actors to other actors and lines. MHA-JAM \cite{messaoud2020multi} concatenates agent features to a CNN-encoded map at their specific coordinates, and then applies attention on this joint representation. The work of \cite{luo2020probabilistic} also uses attention between agents for interactions, and parallely applies an attention head on encoded lane to obtain lane probabilities and generate a modality for each given lane. mmTransformer \cite{liu2021multimodal} applies a general Transformer \cite{vaswani2017polosukhin} architecture to fuse history, map and interactions.

Another family of methods use a pool of anchor trajectories, predefined \cite{chai2020multipath} or model-based \cite{phan2020covernet,song2021learning}, and rank them with a learned model. This allows to avoid any mode collapse and assert realistic trajectories, but removes the ability to tune the trajectories accurately to the current situation.

Multimodality can also be obtained using generative approaches   that model the actual future probability distribution \cite{lee2017desire, mangalam2020not, tang2019multiple, rhinehart2018r2p2, rhinehart2019precog}. However, generative models require multiple independent sampling at inference time without any optimization of coverage or average distance.

More recently, methods have started to leverage the graph obtained from HD-map  in order to better represent lane connectivity. VectorNet \cite{gao2020vectornet} encodes both map features and agent trajectories as polylines then merge them with a global interaction graph. LaneGCN \cite{liang2020learning} treats actor past and the lane graph separately, and then fuse them with a series of attention layers between lane and actors.

Other methods then use the graph to structure their multimodal outputs. TNT \cite{zhao2020tnt} builds from the VectorNet backbone and combines it with multiple target proposals sampled from the lanes in order to diversify the prediction points. GoalNet \cite{zhang2020map} also identifies possible goals and applied a prediction head for each on a localized raster in order to base the modalities on reachable lanes. WIMP \cite{khandelwal2020if} matches possible polylines to the past trajectory and uses them as conditional input to their model. LaneRCNN \cite{zeng2021lanercnn} adds actor features from the start to sampled nodes on the lanes, and then predicts a future point for each node along a probability. 

Grid-based outputs have already been used in pedestrian behavior prediction such as \cite{liang2020garden, deo2020trajectory, mangalam2020goals, jain2020discrete, ridel2020scene}. Their model architecture, training and sampling strategies however differ greatly from ours. The work of \cite{sadat2020perceive} produces a future grid occupancy output prediction for each vehicle class in order to plan from it, but it is not instance-based and doesn't allow for individual vehicle prediction.

\begin{figure*}[t]
\centerline{\includegraphics[width=2\columnwidth]{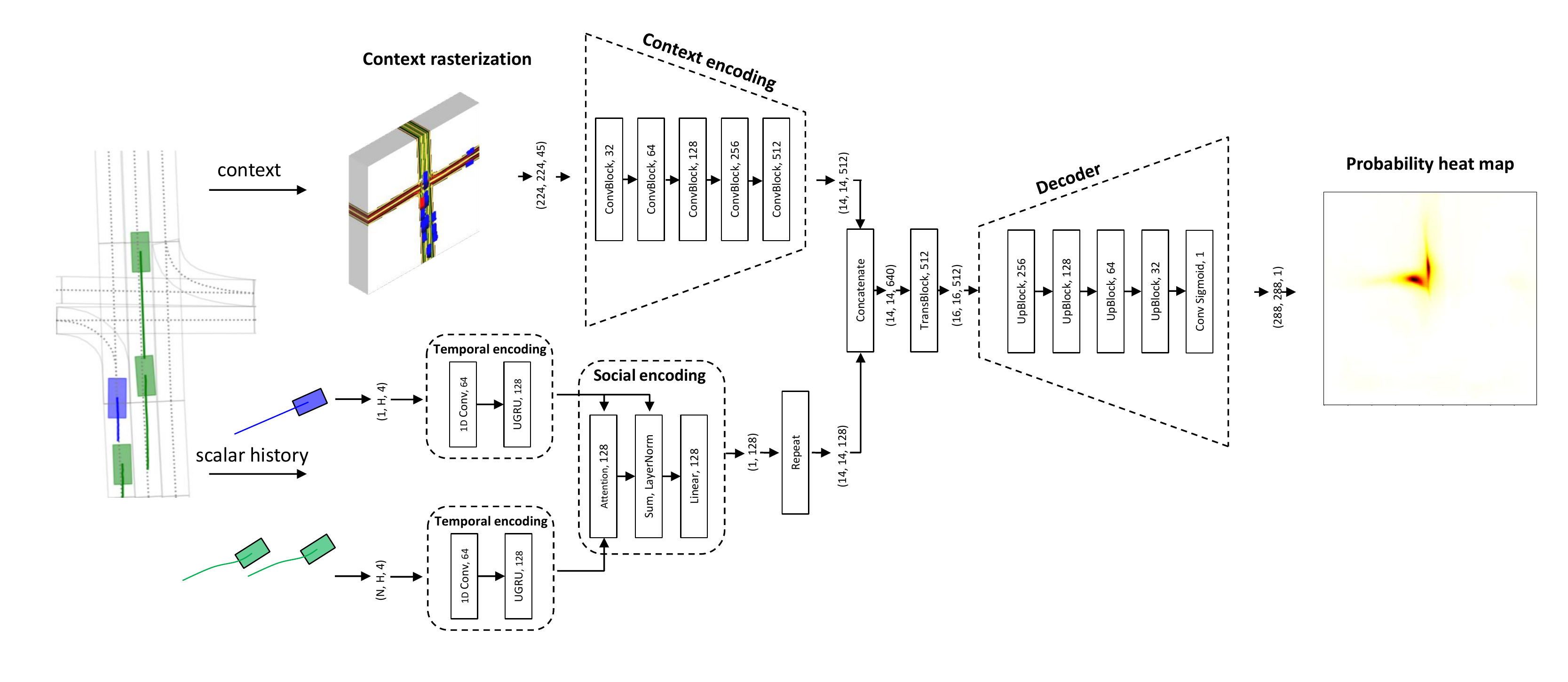}}
\caption{Example of input and output data for our model with brief description of architecture}
\label{fig:in_out_archi}
\end{figure*}

\section{METHOD}

We describe our general pipeline in Fig. \ref{fig:pipeline}. Our method takes as input a rasterized image of the agent environment, and outputs a probability distribution heatmap representing where the agent could be at a fixed time horizon T in the future. A finite set of possible locations are then extracted from the heatmap to ensure appropriate coverage. Future locations are sampled to minimize either rate of misses or final displacement errors. Finally for each sampled future location, a trajectory representing the motion of agent from the initial state to the future location is computed.

The aim of motion estimation is to predict the future positions of the target agent $a$ for T timesteps $\{(x_a^t,y_a^t) \text{ for t in }[\![1, T]\!]\}$. The model is given the past H timesteps $\{(x_a^t,y_a^t) \text{ for t in }[\![-H, 0]\!]\}$ for the target agent $a$ and the $N$ neighbor agents $a'$.
Supplementary context informations are available in the shape of a graph High Definition Map (HD map). We will focus in this paper on the prediction of the final points $(x_a^T,y_a^T)$, and then regress the whole trajectory conditioned on the end point.

\subsection{Encoding history and local context information}

\subsubsection{Map and past trajectory encoding}

The local context is available as a High Definition Map centered on the target agent.
We rasterize the HD-Map in 5 semantic channels: drivable area, lane boundaries and directed center-lines with their headings encoded using HSV on 3 channels. We also add the target agent trajectory as a moving rectangle on 20 history channels and the other agents history on 20 more channels. The final input is a (224, 224, 45) image with a 0.5 x 0.5 m² resolution per pixel. This image is processed by a classic CNN model alternating convolutional layers and max-pooling for downscaling to obtain a (14, 14, 512) encoding $E_{raster}$ as illustrated in the top-left part of Fig. \ref{fig:in_out_archi}.

The scalar history of the agents is also taken as input to the model as a list of 2D coordinates. Missing timesteps are padded with zeros and a binary mask indicating if padding was applied is concatenated to the trajectory, as well as the timestamps for each step, so to obtain a $(H, 4)$ input for each agent. Each agent  trajectory goes through a 1D convolutional layer followed by a UGRU\cite{ugru} recurrent layer. The weights are shared for all agents except the target agent.

\subsubsection{Inter-agent attention for interaction}

Similar to \cite{mercat2020multi, messaoud2020multi, luo2020probabilistic}, we use attention \cite{vaswani2017polosukhin} to model agent interaction. A query vector is generated for the target agent, while key and value vectors are created for the other actors. The normalized dot product of query and keys creates an attention map from the target agent to the other agent, then used to pool their value features into a context vector. The context vector is then added to the target vehicle feature vector through a residual connection followed by LayerNormalization \cite{ba2016layer}. The obtained trajectory encoding $E_{trajectory}$ is then repeated to match the context encoding $E_{raster}$ dimensions. The final encoding $E_{context}$ is the result of the concatenation of both encodings $E_{raster}$ and $E_{trajectory}$.

\subsubsection{Increased output size for longer range}
Due to high speed, some cars may go through a greater range in the time horizon $T$ that is covered by the input range of 56m. However, simply increasing input size would greatly add to the computational burden while not necessarily bringing useful information. We therefore want to increase the output size while retaining the spatial correspondences through the layers. In order to do so, we apply Tranpose Convolutions with stride 1 and kernel size 3. Since 1 input pixel is connected to a grid of 3x3 output pixels, the edge pixels generate a new border of pixels around them, increasing the encoding size by 1 in each direction. We apply 2 of these layers, resulting in a (18, 18, 512) augmented encoding so that once upscaled the decoded image output will be of size (288, 288), corresponding to a 72m range.

\subsection{Heatmap output}

The final part of the model is a convolutional decoder alternating transpose convolutions for upscaling and classic convolutions, topped with a sigmoid activation. We output an image $\hat{Y}$ with similar resolution as the raster input (0.5 x 0.5 m² / pixel). The output target is an image $Y$ with a Gaussian centered around the ground truth position. This image is trained with a pixel-wise focal loss inspired from \cite{zhou2019objects}, averaged over the total $P$ pixels $p$ of the heatmap:

\begin{equation}
\begin{split}
L = -\frac{1}{P}\sum_p (Y_p-\hat{Y}_p)^2 f(Y_p,\hat{Y}_p) \\
\text{with } f(Y_p,\hat{Y}_p)=
    \begin{cases}
      \log(\hat{Y}_p) & \text{if $Y_p$=1}\\
      (1-Y_p)^4\log(1-\hat{Y}_p) & \text{else}
    \end{cases}  
\end{split}
\end{equation}

where the non-null pixels around the Gaussian center serve as penalty-reducing coefficients, and the square factor of error allows the gradient to focus on poorly-predicted pixels. We use a standard deviation of 4 pixels for the Gaussian.

\subsection{Modality sampling}

Our aim is here to sample the probability heatmap in order to optimize the performance metric of our choice. In most datasets such as Argoverse \cite{chang2019argoverse} and NuScenes \cite{caesar2020nuscenes}, two main metrics are used for the final predicted point: MissRate (MR) and Final Displacement Error (FDE). MissRate corresponds to the percentage of prediction being farther than a certain threshhold to the ground truth, and FDE is simply the mean of $l_2$ distance between the prediction and the ground truth.
When the output is multimodal, with $k$ predictions, minimum Final Displacement Error minFDE$_k$ and Miss Rate over the $k$ predictions MR$_k$ are used.

\subsubsection{Optimizing Miss Rate}

We design a sampling method in order to optimize the Miss Rate between the predicted modalities and the ground truth. A case is defined as missed if the ground truth is further than 2m from the prediction. For a given area $A$, the probability of the ground truth $Y$ being in this area is equal to the integral of the probability distribution $p$ under this ground truth. 

\begin{equation}
    P(Y \in A) = \int_{x \in A} p(x) dx
\label{eq:proba}
\end{equation}

Therefore, for $k$ predictions, given a 2D probability distribution, the sampling minimizing the expected MR is the one maximizing the integral of the future probability distribution under the area defined as 2m radius circles around the $k$ predictions:

\begin{equation}
\begin{split}
    E(\mathds{1}_{\min_k\rVert c_k-Y\rVert > 2}) = 1- \sum_k \int_{\rVert c_k-x\rVert < 2} p(x)dx 
\end{split}
\label{eq:mr}
\end{equation}

\begin{figure}[t]
    \center
    \begin{subfigure}[b]{0.23\textwidth}
        \includegraphics[width=\textwidth]{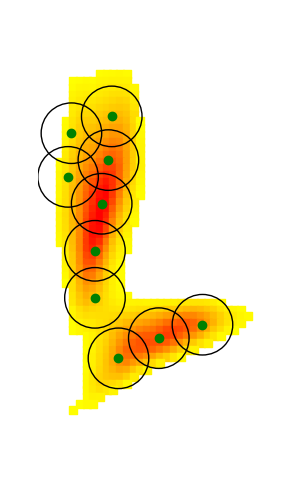}
        \caption{MR sampling}
        \label{fig:mr_samplimg}
    \end{subfigure}
    \begin{subfigure}[b]{0.23\textwidth}
        \includegraphics[width=\textwidth]{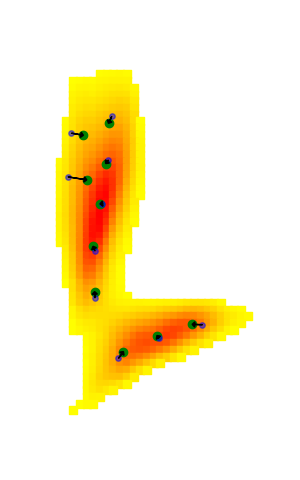}
        \caption{FDE sampling}
        \label{fig:fde_sampling}
    \end{subfigure}
    \caption{Illustration of sampling methods}
    \label{fig:sampling}
\end{figure}

We therefore process in a greedy way as described in Algo. \ref{mr_algo}, and iteratively select the location with the highest integrated probability value in its 2m circle. Once we obtain such a point, we set to zero the heatmap values under the defined circle and move on to selecting the next point with the same method. 

\begin{algorithm}
\DontPrintSemicolon 
\SetKwInOut{Input}{input}
\Input{Probability map p(x) \\

$K$ number of predictions \\

$R$ threshhold for Miss Rate
}
\For{k = 1..K } {
    Find $c_k$ maximizing $ \int_{\rVert c_k-x\rVert < R} p(x)dx $ \\
    Set $p(x)=0$ for all $x$ such that $\rVert c_k-x\rVert < R$ \\
}
\caption{ MR Sampling Algorithm\label{IR}}
\label{mr_algo}
\end{algorithm}

The result is illustrated in Fig. \ref{fig:mr_samplimg}. We see that each sampled point can be surrounded by a circle of radius 2m that barely overlaps with other circles. Each point is sampled almost equidistant to the others, as setting the probability under previous points to zero sets a very strict limit to the minimum distance between points.

For implementation, we process the covered area for each point using a convolution layer with kernel weights fixed so to approximate a 2m circle. In practice, we don't actually use a radius of 2 meters, but a 1.8 meters one as we found out it to yield better performance. We also upscale the heatmap to 0.25 x 0.25 m$_2$ per pixel with bilinear interpolation to have a more refined prediction location.

\subsubsection{Optimizing Final Displacement Error}

We inspire ourselves from KMeans to optimize minFDE$k$.
The image output can be represented as a discrete probability distribution $(x_i, p_i)$ where $x_i$ represents the pixel centers and $p_i$ the associated probability value. Optimizing the Final Displacement Error over $k$ predictions means finding $k$ centroids $c_k$ that minimize the following quantity:

\begin{equation}
    minimize_c \sum_i p_i\rVert c-x_i\rVert
\label{eq:fde}
\end{equation}

To do so we design our sampling algorithm for FDE optimization detailed in Algo. \ref{fde_algo}.

\begin{algorithm}
\DontPrintSemicolon 
\SetKwInOut{Input}{input}
\Input{Set of points $x_i$ with probability weight $p_i$\\

$L$ number of iterations to run the algorithm\\

Initialization of $K$ centroids $c_k$
}
\For{l = 1..L } {
    Compute $d_i^k$ the matrix of distance of point $x_i$ to each centroid $c_k$ \\
    Compute $m_i$ the distance of point $x_i$ to the closest centroid $c_k$ \\
    \For{k = 1..K } {
        Compute new centroid coordinates :
        $$ c_k = \frac{1}{N} \sum_i \mathds{1}_{d_i^k<=3} \frac{p_i}{d_i^k} \frac{m_i}{d_i^k} x_i $$
        with $N = \sum_i \mathds{1}_{d_i^k<=3} \frac{p_i}{d_i^k} \frac{m_i}{d_i^k}$

    }
}
\caption{ FDE Sampling Algorithm\label{IR}}
\label{fde_algo}
\end{algorithm}

We replace the classic weighted average $\sum_i p_i x_i$ for each centroid $c_k$ by $\sum_i \frac{p_i}{d^k_i} x_i$ where $d^k_i$ is the distance between point $x_i$ and centroid $c_k$ to be more robust to outliers and take into account the optimisation of $l_2$ norm instead of its square.

In essence, we update each prediction as a weighted average of its local neighborhood in a radius of 3m. The coefficient $\frac{m_i}{d_i^k}$, with $m_i$ the distance between point $x_i$ and its closest centroid allows for flexible partition boundaries compared to KMeans (where we would use $\mathds{1}_{d_i^k<=m_i}$ instead): when $x_i$ is in the partition of prediction $k$, its value is 1, while when it's outside it decreases, so as to be 0 when at the exact position of another prediction $k'$, where it could never be improved by a displacement of $k$.

We initialize the centroids with the results of the Miss Rate optimization algorithm and use the number of iterations $L$ as a parameter to tune the trade-off between Miss Rate and FDE: when $L$ is zero, Miss Rate is optimized while when $L$ increases MR is sacrificed to get better FDE. The output of the algorithm is illustrated in Fig. \ref{fig:fde_sampling}, where is it can be observed that centroids are brought closer together, sacrificing total coverage but getting closer to areas with high probabilities to reduce the expected distance. Results of this trade-off are illustrated further in Sec. \ref{sec:traj_trade}, where we show in Fig. \ref{fig:trade} that every iteration of Algo. \ref{fde_algo} diminishes minFDE$_6$ and increases MR$_6$.

\begin{table*}[bp]
\caption{Results on Argoverse Motion Forecasting Leaderboard \cite{leaderboard} (test set)}
    \begin{center}
    \begin{tabular}{l|c c c|c c c c a}
      \hline
       & \multicolumn{3}{c|}{K=1}  & \multicolumn{5}{c}{K=6}  \\
      & minADE & minFDE& MR&minADE & minFDE&p-minADE & p-minFDE& MR \\
      \hline
      WIMP \cite{khandelwal2020if} & 1.82 & 4.03 & 62.9 &  0.90 & 1.42& 2.69 & 3.21 & 16.7\\
      LaneGCN \cite{liang2020learning} & 1.71 & 3.78 & 59.1 &  0.87 & 1.36 & 2.66 & 3.16 & 16.3\\
      Alibaba-ADLab  & 1.97 & 4.35 & 63.4 &  0.92 & 1.48 & 2.64 & 3.23 & 15.9\\
      TPCN \cite{ye2021tpcn} & \textbf{1.64} & \textbf{3.64} & 58.6 &  \textbf{0.85} & \textbf{1.35} & 2.61 & 3.11 & 15.9\\
      HIKVISION-ADLab-HZ & 1.94 & 3.90 & 58.2 &  1.21 & 1.83 & 3.00 & 3.62& 13.8\\
      TNT \cite{zhao2020tnt} & 1.78 & 3.91 & 59.7 &  0.94 & 1.54 & 2.73 & 3.33& 13.3\\
      Jean \cite{mercat2020multi} & 1.74 & 4.24 & 68.6 &  1.00 & 1.42 & 2.79 & 3.21 & 13.1\\
      TMP \cite{liu2021multimodal} & 1.70 & 3.78 & 58.4 & 0.87 & 1.37 & 2.66 & 3.16 & 13.0\\
      LaneRCNN \cite{zeng2021lanercnn} & 1.69 & 3.69 & \textbf{56.9} &  0.90 & 1.45 & 2.70 & 3.24 & 12.3\\
      SenseTime\_AP  & 1.70 & 3.76 & 58.3 &  0.87 & 1.36 & 2.66 & 3.16 & 12.0\\
      poly (3$^{rd}$) & 1.70 & 3.82 & 58.8 &  0.87 & 1.47 & 2.67 & 3.28 & 12.0\\
      PRIME (2$^{nd}$) \cite{song2021learning}& 1.91 & 3.82 & 58.7 &  1.22 & 1.56 & 2.71 & 3.04 & 11.5\\
      \hline
      Ours-HOME (FDE L=4) & 1.72 & 3.73 & 58.4 &  0.92 & 1.36 & 2.64 & 3.08 & 11.3\\
      Ours-HOME (MR) (1$^{st}$) & 1.73 & 3.73 & 58.4 &  0.94 & 1.45 & \textbf{2.52} & \textbf{3.03}& \textbf{10.2}\\

      \hline
    \end{tabular}
    \end{center}
    \label{tab:argo_test}
\end{table*}

\subsection{Full trajectory generation}

We use a separate model to generate full trajectories connecting the initial agent position to all sampled locations. This model applies a fully-connected layer to encode the target agent history into a vector of 32 features, which is then concatenated with the $(x,y)$ coordinates of the target future location. Another fully-connected layer is then applied to obtain a 64 feature vector, which is then transformed through a last fully-connected layer to a set of locations representing the intermediate position of the agent in the time frame $[\![1, T]\!]$. The probability of a trajectory is the integral of the probability heatmap under the circle of radius 2m around the end point of the trajectory.

\section{EXPERIMENTS}

\subsection{Experimental settings}

\subsubsection{Dataset} We use the Argoverse Motion Forecasting Dataset \cite{chang2019argoverse}. It is a car trajectory prediction benchmark with 205942 training samples, 39472 validation samples and 78143 test samples. Each sample contains the position of all agents in the scene in the past 2s as well as the local map, and the labels are the 3s future positions of one target agent in the scene. 

\subsubsection{Metrics}

We report the previously defined metrics MR$_k$ and minFDE$_k$ for k=1,6, completed by the minimum Average Displacement Error minADE$_k$ which is the average $l_2$ error over all successive trajectory points. We also report the metrics p-minFDE$_6$ and p-minADE$_6$ for the test set, where $-\log(p)$ is added to the metric, $p$ being the probability assigned to the best (closest to ground-truth) predicted trajectory. These later metrics allow to measure the quality of the probability distribution assigned to the predictions.

\subsubsection{Implementation details} We train all models for 16 epochs with batch size 32, using Adam optimizer initialized with a learning rate of 0.001. Each sample frame is centered on the target agent and aligned with its heading. We divide learning rate by half at epochs 3, 6, 9 and 13. We augment the training data by dropping each raster channel with a probability of 0.1 and rotating the frame by a uniform random angle in $[-\pi/4, \pi/4]$ in 50\% of the samples. All convolution layers are CoordConv \cite{liu2018intriguing} with a kernel of size 3x3 (3 for 1D Convs) and are followed by BatchNormalization and ReLU activation.

\subsection{Comparison with State-of-the-art}

We show in Tab. \ref{tab:argo_test} our results compared to other methods on the Argoverse motion forecasting test set. The benchmark is ranked by MR$_6$, where we rank first and significantly improve on previous results, demonstrating that having the heatmap output enables the best coverage with respect to the prior art. We also outperform other methods on both p-minFDE$_6$ and p-minADE$_6$, demonstrating superior modelling of the probability distribution between predictions. Another interesting observation is that methods performing very well on minFDE$_6$ such as LaneGCN \cite{liang2020learning} and TPCN \cite{ye2021tpcn} have a worse MR$_6$ as drawback. PRIME \cite{song2021learning} has the closest MR$_6$ to ours but a much higher minFDE$_6$ in comparison. We show the results of both our sampling optimized for MR and minFDE with the same trained model. Our FDE sampling with $L=4$ sacrifices 1.1 points of MR$_6$ for 9 cm of minFDE$_6$, which gets us second best on minFDE$_6$ while still being good enough for 1$^{st}$ position on the leaderboard.

\begin{table}[t]
\centering
\caption{Ablation study on output representation \\(Argoverse validation set)}
\begin{tabular}{l|c|c c|c c}
  \hline
   Bottleneck & Output & \multicolumn{2}{c|}{K=1}  & \multicolumn{2}{c}{K=6}  \\
  & & minFDE& MR& minFDE& MR \\
  \hline
  Scalar & Regression & 3.81 & 61.7  & 1.26 & 13.0\\
  Scalar & Heatmap  & 3.07 & 51.9 &  1.30 & 8.0\\
  Image & Heatmap   & 3.02 & 50.7 & 1.28 & 6.8\\
  \hline
\end{tabular}
%   \end{center}
\label{tab:ablation_output}
\end{table}

% \begin{table}[t]
% \caption{Ablation study of modules}
%     \begin{center}
%     \begin{tabular}{c c |c c|c c}
%       \hline
%       Scalar history & Vehicle Attention  & \multicolumn{2}{c|}{K=1}  & \multicolumn{2}{c}{K=6}  \\
%       & & minFDE& MR& minFDE& MR \\
%       \hline
%       & & _ & _  & _ & _\\
%       \checkmark &  & _ & _ &  _ & _\\
%       \checkmark & \checkmark   & 3.02 & 50.7 & 1.28 & 6.8\\

%       \hline
%     \end{tabular}
%     \end{center}
%     \label{tab:ablation_archi}
% \end{table}

\subsection{Ablation studies}

We discuss the importance of our difference contributions, starting by comparing our output representation to the traditional scalar coordinates output, then decomposing our model architecture and sampling strategies. All metrics are reported on the Argoverse validation set. If not specified otherwise, MR sampling is used.

\subsubsection{Heatmap output}

We show the effect of output representation in Tab. \ref{tab:ablation_output} by using the same encoding backbone and replacing the image decoder with a global pooling followed by a regression head of 6 coordinate modalities. We train the regression output with a winner-takes-all $l_1$ regression loss similar to \cite{messaoud2020multi, liang2020learning, khandelwal2020if, ye2021tpcn, cui2019multimodal} and a classification loss where target is obtained through a softmax on distances between predictions and ground-truth, as in \cite{zhao2020tnt,song2021learning}. Since the global pooling leads to loss of spatial information from the image, for fair comparison we also include a model with "scalar bottleneck" where pooling is also applied on the image encoding and is then reshaped to form an image on which is applied the heatmap decoder. We observe that heatmap outputs yields much better Miss Rate, and that having a scalar pooling bottleneck diminishes performance as it creates information loss, but not significantly. Interestingly, the regression output reaches better minFDE$_6$ when compared to the MR-optimized sampled image output models, but is still worse than FDE-optimized model, as this scalar coordinates output doesn't leave room for any post-processing optimization.

\begin{figure}[tp]
\centerline{\includegraphics[width=\columnwidth]{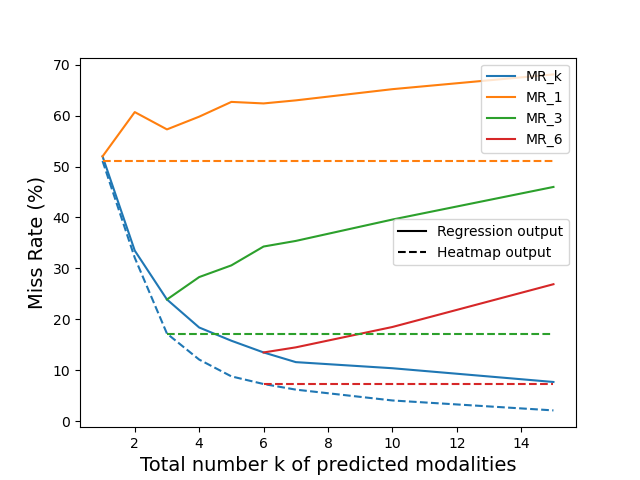}}
\caption{Effect of maximum number $k$ of modalities trained on metrics of lower fixed modality numbers. Full lines are results of regression output model. Dashed lines are result of our heatmap output model. We show the Miss Rate for total number of predicted modalities $k$ (blue) and fixed number of modalities 1 (orange), 3 (green) and 6 (red).}
\label{fig:increase}
\end{figure}

We also show the effect of adding more modalities to a regression output in Fig \ref{fig:increase} : even if the MR$_k$ improves for the total number of modalities as $k$ increases, the performance for a fixed $k$ such as 1 or 6 worsens. \cite{khandelwal2020if} and \cite{zhang2020map} notice a similar trend, obtaining much better results for lower $k$ metrics when training less modalities. Furthermore, for a regression output model a new training is required each time to accommodate the maximum number of modalities, whereas with heatmap output any number of modalities can be obtained at will with the same training, and the lower k numbers are not impacted by the total number of modalities extracted, as showed by the dashed horizontal lines displayed for MR$_1$, MR$_3$ and MR$_6$. Finally, our model heatmap output scales better with the number of $k$ modalities, converging to a 0\% MR faster that the regression output model.

% \subsubsection{Architecture}

% We study the effect of adding the scalar coordinates history of the target agent as input to the model in addition to its rasterized context, and check that the attention module allows for better understanding of interactions leading to better performance. We report the results in Tab. \ref{tab:ablation_archi}. Having the scalar history brings significant improvement as it allows the model to have sub-pixel sensitivity, and attention towards other agent enables better interaction understand, leading to a better score as well.

\subsubsection{Trajectory sampling}
\label{sec:traj_trade}

\begin{figure}[bp]
\centerline{\includegraphics[width=\columnwidth]{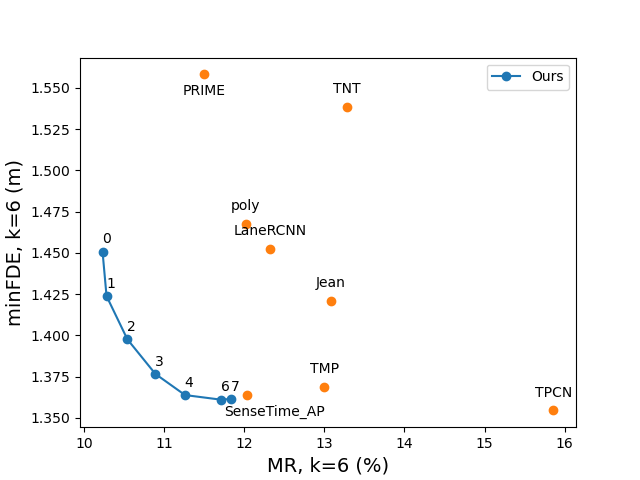}}
\caption{FDE$_6$ - MR$_6$ trade-off. Lower-left is better. Points of the curve (blue) are obtained increasing number of iteration $L$ of Algorithm \ref{fde_algo} from 0 to 7. Points for other top-10 leaderboard methods are also included (orange).}
\label{fig:trade}
\end{figure}

We show in Fig. \ref{fig:trade} the results of our trade-off between MR$_6$ and FDE$_6$ on the Argoverse test set  thanks to the parameter $L$ of Algo. \ref{fde_algo}. We also include points for the other top 10 methods of the leaderboard for comparison. Our method reaches best possible MR$_6$, and allows to improve FDE$_6$ to second-best while still being first in MR$_6$ (fourth curve point obtained with $L=4$)

\begin{table}[t]
\caption{Ablation study on trajectory sampling \\(Argoverse validation set)}
    \begin{center}
    \begin{tabular}{l|c c|c c}
      \hline
       Bottleneck  & \multicolumn{2}{c|}{K=1}  & \multicolumn{2}{c}{K=6}  \\
      & minFDE& MR& minFDE& MR \\
      \hline
      Pixel ranking with NMS & 3.07 & 51.0  & 1.21 & 10.7\\
      KMeans  & 3.06 & 51.6 &  1.23 & 9.3\\
      Ours (MR)   & 3.02 & 50.7 & 1.28 & \textbf{6.8}\\
      Ours (FDE L=6)   & \textbf{3.01} & \textbf{50.5} & \textbf{1.16} & 7.4\\

      \hline
    \end{tabular}
    \end{center}
    \label{tab:ablation_sampling}
\end{table}

We highlight our sampling results in Tab \ref{tab:ablation_sampling} and compare them to other possible sampling strategies: we try ranking pixels by probability and select them in decreasing order while removing overlapping pixels that are closer than a 1.8m radius following a classic Non-Maximum Suppression method. We also try KMeans as is used in \cite{mangalam2020goals}.

%\subsection{Results on other datasets}

\begin{figure*}[b]
\centerline{\includegraphics[width=2\columnwidth]{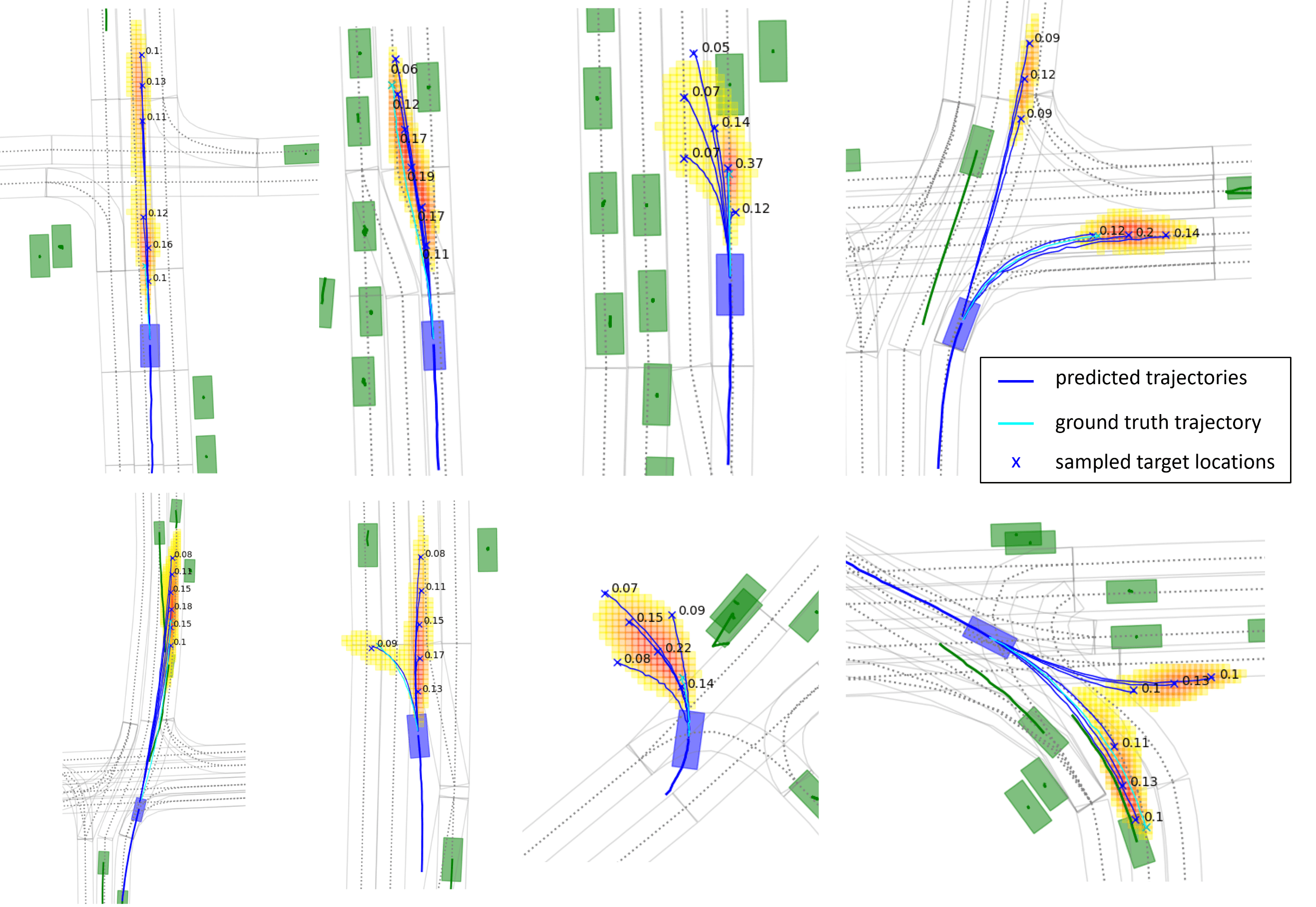}}
\caption{Qualitative examples. The yellow/red heatmap is our predicted probability distribution and the blue points are the sampled final point predictions. The ground truth trajectory is shown in green.}
\label{fig:qualitative}
\end{figure*}

\subsection{Qualitative results}

We show supplementary qualitative results in Fig. \ref{fig:qualitative}. We highlight examples of straight line, overtaking, curve road, going outside the map and intersections. Our model heatmap output makes use and usually follows the prior from the context map, but it is also able to divert from it based on interactions, realistic observations and hints of divergence from history.

\section{CONCLUSION}

We have presented HOME, a novel representation for multimodal trajectory prediction. It is based on predicting the future final point position on a 2D top-view grid, decoding then this final point into a full trajectory. This heatmap output represents the complete future probability distribution and its uncertainties, from which we design two prediction sampling methods. Sampling directly from the heatmap distribution enables a more optimized coverage, achieving state-of-the-art performance on the Argoverse Motion Forecasting benchmark.

%\addtolength{\textheight}{-12cm}   % This command serves to balance the column lengths
                                  % on the last page of the document manually. It shortens
                                  % the textheight of the last page by a suitable amount.
                                  % This command does not take effect until the next page
                                  % so it should come on the page before the last. Make
                                  % sure that you do not shorten the textheight too much.

%%%%%%%%%%%%%%%%%%%%%%%%%%%%%%%%%%%%%%%%%%%%%%%%%%%%%%%%%%%%%%%%%%%%%%%%%%%%%%%%

%%%%%%%%%%%%%%%%%%%%%%%%%%%%%%%%%%%%%%%%%%%%%%%%%%%%%%%%%%%%%%%%%%%%%%%%%%%%%%%%

%%%%%%%%%%%%%%%%%%%%%%%%%%%%%%%%%%%%%%%%%%%%%%%%%%%%%%%%%%%%%%%%%%%%%%%%%%%%%%%%

\section*{ACKNOWLEDGMENT}

We would like to thank Thomas Wang and Camille Truong-Allié for useful comments on the paper, as well as Arthur Moreau and Joseph Gesnouin for insightful discussions.

%%%%%%%%%%%%%%%%%%%%%%%%%%%%%%%%%%%%%%%%%%%%%%%%%%%%%%%%%%%%%%%%%%%%%%%%%%%%%%%%

%\addtolength{\textheight}{-16cm}

%\bibliographystyle{IEEEtran}
%\bibliography{references}

\printbibliography

@inproceedings{gao2020vectornet,
  title={Vectornet: Encoding hd maps and agent dynamics from vectorized representation},
  author={Gao, Jiyang and Sun, Chen and Zhao, Hang and Shen, Yi and Anguelov, Dragomir and Li, Congcong and Schmid, Cordelia},
  booktitle={CVPR},
%   booktitle={Proceedings of the IEEE/CVF Conference on Computer Vision and Pattern Recognition}, 
%   pages={11525--11533},
  year={2020}
}

@inproceedings{liang2020learning,
  title={Learning lane graph representations for motion forecasting},
  author={Liang, Ming and Yang, Bin and Hu, Rui and Chen, Yun and Liao, Renjie and Feng, Song and Urtasun, Raquel},
  booktitle={ECCV},
%   booktitle={European Conference on Computer Vision},
%   pages={541--556},
  year={2020},
  %organization={Springer}
}

@article{zhao2020tnt,
  title={TNT: Target-driven trajectory prediction},
  author={Zhao, Hang and Gao, Jiyang and Lan, Tian and Sun, Chen and Sapp, Benjamin and Varadarajan, Balakrishnan and Shen, Yue and Shen, Yi and Chai, Yuning and Schmid, Cordelia and others},
  %journal={arXiv preprint arXiv:2008.08294},
  %journal={arXiv:2008.08294},
  journal={CoRL},
  year={2020}
}

@inproceedings {zhang2020map,
  title={Map-Adaptive Goal-Based Trajectory Prediction},
  author={Zhang, Lingyao and Su, Po-Hsun and Hoang, Jerrick and Haynes, Galen Clark and Marchetti-Bowick, Micol},
  %journal={arXiv preprint arXiv:2009.04450},
  booktitle={CoRL},
  year={2020}
}

@article{zeng2021lanercnn,
  title={LaneRCNN: Distributed Representations for Graph-Centric Motion Forecasting},
  author={Zeng, Wenyuan and Liang, Ming and Liao, Renjie and Urtasun, Raquel},
  %journal={arXiv preprint arXiv:2101.06653},
  journal={arXiv:2101.06653},
  year={2021}
}

@inproceedings{mercat2020multi,
  title={Multi-head attention for multi-modal joint vehicle motion forecasting},
  author={Mercat, Jean and Gilles, Thomas and El Zoghby, Nicole and Sandou, Guillaume and Beauvois, Dominique and Gil, Guillermo Pita},
%   booktitle={2020 IEEE International Conference on Robotics and Automation (ICRA)},
%   pages={9638--9644},
  booktitle = {ICRA},
  year={2020},
  %organization={IEEE}
}

@article{messaoud2020multi,
  title={Multi-Head Attention with Joint Agent-Map Representation for Trajectory Prediction in Autonomous Driving},
  author={Messaoud, Kaouther and Deo, Nachiket and Trivedi, Mohan M and Nashashibi, Fawzi},
  journal={arXiv:2005.02545},
  %journal={arXiv preprint arXiv:2005.02545},
  year={2020}
}

@article{luo2020probabilistic,
  title={Probabilistic Multi-modal Trajectory Prediction with Lane Attention for Autonomous Vehicles},
  author={Luo, Chenxu and Sun, Lin and Dabiri, Dariush and Yuille, Alan},
  %journal={arXiv preprint arXiv:2007.02574},
  journal={arXiv:2007.02574},
  year={2020}
}

@article{zhou2019objects,
  title={Objects as points},
  author={Zhou, Xingyi and Wang, Dequan and Kr{\"a}henb{\"u}hl, Philipp},
  %journal={arXiv preprint arXiv:1904.07850},
  journal={arXiv:1904.07850},
  year={2019}
}

@inproceedings{chang2019argoverse,
  title={Argoverse: 3d tracking and forecasting with rich maps},
  author={Chang, Ming-Fang and Lambert, John and Sangkloy, Patsorn and Singh, Jagjeet and Bak, Slawomir and Hartnett, Andrew and Wang, De and Carr, Peter and Lucey, Simon and Ramanan, Deva and others},
%   booktitle={Proceedings of the IEEE/CVF Conference on Computer Vision and Pattern Recognition},
%   pages={8748--8757},
  booktitle={CVPR},
  year={2019}
}

@inproceedings{caesar2020nuscenes,
  title={nuScenes: A multimodal dataset for autonomous driving},
  author={Caesar, Holger and Bankiti, Varun and Lang, Alex H and Vora, Sourabh and Liong, Venice Erin and Xu, Qiang and Krishnan, Anush and Pan, Yu and Baldan, Giancarlo and Beijbom, Oscar},
%   booktitle={Proceedings of the IEEE/CVF conference on computer vision and pattern recognition},
%   pages={11621--11631},
  booktitle={CVPR},
  year={2020}
}

@article{ye2021tpcn,
  title={TPCN: Temporal Point Cloud Networks for Motion Forecasting},
  author={Ye, Maosheng and Cao, Tongyi and Chen, Qifeng},
%   journal={arXiv preprint arXiv:2103.03067},
  journal={arXiv:2103.03067},
  year={2021}
}

@article{khandelwal2020if,
  title={What-If Motion Prediction for Autonomous Driving},
  author={Khandelwal, Siddhesh and Qi, William and Singh, Jagjeet and Hartnett, Andrew and Ramanan, Deva},
%   journal={arXiv preprint arXiv:2008.10587},
  journal={arXiv:2008.10587},
  year={2020}
}

@inproceedings{cui2019multimodal,
  title={Multimodal trajectory predictions for autonomous driving using deep convolutional networks},
  author={Cui, Henggang and Radosavljevic, Vladan and Chou, Fang-Chieh and Lin, Tsung-Han and Nguyen, Thi and Huang, Tzu-Kuo and Schneider, Jeff and Djuric, Nemanja},
%   booktitle={2019 International Conference on Robotics and Automation (ICRA)},
%   pages={2090--2096},
  booktitle={ICRA},
  year={2019},
%  organization={IEEE}
}

@article{song2021learning,
  title={Learning to Predict Vehicle Trajectories with Model-based Planning}, 
  author={Haoran Song and Di Luan and Wenchao Ding and Michael Yu Wang and Qifeng Chen},
  journal={arXiv:2103.04027},
  year={2021},
    %   eprint={2103.04027},
    %   archivePrefix={arXiv},
    %   primaryClass={cs.CV}
}

@misc{leaderboard,
  title = {Argoverse motion forecasting competition},
  howpublished = {\url{https://eval.ai/web/challenges/challenge-page/454/leaderboard/1279}},
  note = {Accessed: 2021-03-12}
}

@inproceedings{chai2020multipath,
  title={MultiPath: Multiple Probabilistic Anchor Trajectory Hypotheses for Behavior Prediction},
  author={Chai, Yuning and Sapp, Benjamin and Bansal, Mayank and Anguelov, Dragomir},
  booktitle={CoRL},
%   booktitle={Conference on Robot Learning},
%   pages={86--99},
  year={2020},
%  organization={PMLR}
}

@inproceedings{phan2020covernet,
  title={Covernet: Multimodal behavior prediction using trajectory sets},
  author={Phan-Minh, Tung and Grigore, Elena Corina and Boulton, Freddy A and Beijbom, Oscar and Wolff, Eric M},
  booktitle={CVPR},
%   booktitle={Proceedings of the IEEE/CVF Conference on Computer Vision and Pattern Recognition},
%   pages={14074--14083},
  year={2020}
}

@inproceedings{liang2020garden,
  title={The garden of forking paths: Towards multi-future trajectory prediction},
  author={Liang, Junwei and Jiang, Lu and Murphy, Kevin and Yu, Ting and Hauptmann, Alexander},
  booktitle={CVPR},
%   booktitle={Proceedings of the IEEE/CVF Conference on Computer Vision and Pattern Recognition},
%   pages={10508--10518},
  year={2020}
}

@article{deo2020trajectory,
  title={Trajectory forecasts in unknown environments conditioned on grid-based plans},
  author={Deo, Nachiket and Trivedi, Mohan M},
  %journal={arXiv preprint arXiv:2001.00735},
  journal={arXiv:2001.00735},
  year={2020}
}

@article{mangalam2020goals,
  title={From Goals, Waypoints \& Paths To Long Term Human Trajectory Forecasting},
  author={Mangalam, Karttikeya and An, Yang and Girase, Harshayu and Malik, Jitendra},
  %journal={arXiv preprint arXiv:2012.01526},
  journal={arXiv:2012.01526},
  year={2020}
}

@inproceedings{vaswani2017polosukhin,
  title={Attention is all you need},
  author={Vaswani, A and Shazeer, N and Parmar, N and Uszkoreit, J and Jones, L and Gomez, AN},
  booktitle={NIPS},
%   journal={Advances in neural information processing systems},
%   pages={5998--6008},
  year={2017}
}

@inproceedings{ba2016layer,
  title={Layer normalization},
  author={Ba, Jimmy Lei and Kiros, Jamie Ryan and Hinton, Geoffrey E},
  %booktitle={arXiv preprint arXiv:1607.06450},
  booktitle={arXiv:1607.06450},
  year={2016}
}

@inproceedings{mangalam2020not,
  title={It is not the journey but the destination: Endpoint conditioned trajectory prediction},
  author={Mangalam, Karttikeya and Girase, Harshayu and Agarwal, Shreyas and Lee, Kuan-Hui and Adeli, Ehsan and Malik, Jitendra and Gaidon, Adrien},
  booktitle={ECCV},
%   booktitle={European Conference on Computer Vision},
%   pages={759--776},
  year={2020},
  %organization={Springer}
}

@inproceedings{jain2020discrete,
  title={Discrete residual flow for probabilistic pedestrian behavior prediction},
  author={Jain, Ajay and Casas, Sergio and Liao, Renjie and Xiong, Yuwen and Feng, Song and Segal, Sean and Urtasun, Raquel},
%   booktitle={Conference on Robot Learning},
  booktitle={ECCV},
  %pages={407--419},
  year={2020},
  %organization={PMLR}
}

@inproceedings{sadat2020perceive,
  title={Perceive, predict, and plan: Safe motion planning through interpretable semantic representations},
  author={Sadat, Abbas and Casas, Sergio and Ren, Mengye and Wu, Xinyu and Dhawan, Pranaab and Urtasun, Raquel},
  booktitle={ECCV},
%   booktitle={European Conference on Computer Vision},
%   pages={414--430},
  year={2020},
  %organization={Springer}
}

@article{ridel2020scene,
  title={Scene compliant trajectory forecast with agent-centric spatio-temporal grids},
  author={Ridel, Daniela and Deo, Nachiket and Wolf, Denis and Trivedi, Mohan},
  journal={IEEE Robotics and Automation Letters},
%   volume={5},
%   number={2},
%   pages={2816--2823},
  year={2020},
  %publisher={IEEE}
}

@inproceedings{lee2017desire,
  title={Desire: Distant future prediction in dynamic scenes with interacting agents},
  author={Lee, Namhoon and Choi, Wongun and Vernaza, Paul and Choy, Christopher B and Torr, Philip HS and Chandraker, Manmohan},
  booktitle={CVPR},
%   booktitle={Proceedings of the IEEE Conference on Computer Vision and Pattern Recognition},
%   pages={336--345},
  year={2017}
}

@inproceedings{tang2019multiple,
  title={Multiple Futures Prediction},
  author={Tang, Yichuan Charlie and Salakhutdinov, Ruslan},
  %booktitle={Advances in neural information processing systems},
  booktitle={NeurIPS},
  year={2019}
}

@inproceedings{rhinehart2018r2p2,
  title={R2p2: A reparameterized pushforward policy for diverse, precise generative path forecasting},
  author={Rhinehart, Nicholas and Kitani, Kris M and Vernaza, Paul},
  booktitle={ECCV},
%   booktitle={Proceedings of the European Conference on Computer Vision (ECCV)},
%   pages={772--788},
  year={2018}
}

@inproceedings{rhinehart2019precog,
  title={Precog: Prediction conditioned on goals in visual multi-agent settings},
  author={Rhinehart, Nicholas and McAllister, Rowan and Kitani, Kris and Levine, Sergey},
%   booktitle={Proceedings of the IEEE/CVF International Conference on Computer Vision},
%   pages={2821--2830},
  booktitle={CVPR},
  year={2019}
}

@article{mozaffari2020deep,
  title={Deep learning-based vehicle behavior prediction for autonomous driving applications: A review},
  author={Mozaffari, Sajjad and Al-Jarrah, Omar Y and Dianati, Mehrdad and Jennings, Paul and Mouzakitis, Alexandros},
  journal={IEEE Transactions on Intelligent Transportation Systems},
  year={2020},
  %publisher={IEEE}
}

@misc{ugru,
  title = {6th Place Solution: Very Custom GRU},
  journal={Riiid Answer Correctness Prediction},
  author={Erdem, Ahmet},
  howpublished = {\url{www.kaggle.com/c/riiid-test-answer-prediction/discussion/209581}}
}

@inproceedings{liu2018intriguing,
  title={An intriguing failing of convolutional neural networks and the CoordConv solution},
  author={Liu, Rosanne and Lehman, Joel and Molino, Piero and Such, Felipe Petroski and Frank, Eric and Sergeev, Alex and Yosinski, Jason},
  booktitle={NeurIPS},
%   booktitle={Proceedings of the 32nd International Conference on Neural Information Processing Systems},
%   pages={9628--9639},
  year={2018}
}

@article{liu2021multimodal,
  title={Multimodal Motion Prediction with Stacked Transformers},
  author={Liu, Yicheng and Zhang, Jinghuai and Fang, Liangji and Jiang, Qinhong and Zhou, Bolei},
  journal={arXiv preprint arXiv:2103.11624},
  year={2021}
}

\end{document}